%% file: main.tex
\documentclass[10pt,twocolumn,letterpaper]{article}

\usepackage[pagenumbers]{cvpr}

\input{preamble}

\definecolor{cvprblue}{rgb}{0.21,0.49,0.74}
\usepackage[pagebackref,breaklinks,colorlinks,allcolors=cvprblue]{hyperref}

\def\paperID{46}
\def\confName{CVPR}
\def\confYear{2026}

\title{Scheduled Style Injection: Expanding the Style-Content Pareto Frontier\\
in Training-Free Diffusion-based Style Transfer}

\author{Amey Sunil Kulkarni\\
Independent Researcher\\
{\tt\small amey1695@gmail.com}
}

\begin{document}
\maketitle
\input{sec/0_abstract}
\input{sec/1_intro}
\input{sec/2_related}
\input{sec/3_method}
\input{sec/4_experiments}
\input{sec/5_results}

\input{sec/6_analysis}

\input{sec/7_conclusion}
{
    \small
    \bibliographystyle{ieeenat_fullname}
    \bibliography{main}
}

\appendix
\input{sec/X_suppl}

\end{document}

%% file: preamble.tex
\usepackage{booktabs}    
\usepackage{multirow}    
\usepackage{makecell}    
\usepackage{colortbl}    
\usepackage{xcolor}      
\usepackage{flushend}    
\usepackage{float}       
\usepackage{placeins}    

\newcommand{\gain}[1]{{\,\scriptsize\textcolor{gray}{($-$#1\%)}}}
\newcommand{\loss}[1]{{\,\scriptsize\textcolor{gray}{($+$#1\%)}}}

\newcommand{\TODO}[1]{\textbf{\color{red}[TODO: #1]}}
\renewcommand{\TODO}[1]{}



%% file: sec/0_abstract.tex
\begin{abstract}
Style transfer with pre-trained diffusion models has advanced rapidly, but a core question remains underexplored: where in the model should style injection be strongest?
StyleID, the leading training-free method, uses a single global parameter (gamma) uniformly across all layers and timesteps, which forces a fixed tradeoff between style quality and content preservation.
We show this tradeoff is unnecessarily rigid.
We systematically explore four dimensions of control: varying style injection strength across decoder layers, across denoising timesteps, and scheduling ControlNet geometric conditioning along both axes.
The pattern is consistent everywhere: decreasing schedules, with stronger structural signal injection in shallower layers and earlier timesteps, reliably outperform the reverse.
Beyond direction, schedule shape matters: cosine and square-root timestep schedules outperform linear. Most importantly, we find that gamma scheduling and ControlNet conditioning are nearly independent. The resulting combined configurations expand the Pareto frontier, offering superior tradeoffs between style fidelity and content preservation compared to any single baseline setting.
Our best balanced configuration achieves ArtFID of 27.036 versus StyleID's 28.801 - a 6.1\% relative improvement, with consistent gains across the full style-content tradeoff frontier. Results are validated across 35 configurations totaling over 28,000 stylized images using four complementary metrics. These findings generalize across SD backbones with identical rank ordering. All modifications are training-free, parameter-free, and require only a few lines of scheduling code; code is available at \url{https://github.com/ameyskulkarni/scheduled_style_injection}.
\end{abstract}

%% file: sec/1_intro.tex
\section{Introduction}
\label{sec:intro}

Style transfer asks a model to do two things at once: preserve the content of one image while adopting the visual style of another.
Since Gatys \etal~\cite{gatys2016image} showed that convolutional feature statistics can capture artistic style, the field has progressed through adaptive normalization~\cite{huang2017adain}, attention-based matching~\cite{park2019sanet}, and theoretical grounding establishing style transfer as feature distribution alignment~\cite{li2017demystifying}.
Throughout all of these approaches, the core tension has remained: pushing harder on style fidelity degrades content preservation, and vice versa.

Denoising diffusion probabilistic models~\cite{ho2020ddpm} and, in particular, large-scale latent diffusion models~\cite{rombach2022ldm} have changed the nature of this problem.
These models develop rich internal structure that can be exploited at inference time without retraining.
Self-attention~\cite{vaswani2017attention} layers encode texture and local appearance.
Decoder layers progressively build spatial structure.
The denoising trajectory moves from coarse composition to fine detail~\cite{tumanyan2023plugandplay,hertz2024editfriendly}.
All of this can be steered by injecting or redirecting internal features during denoising~\cite{hertz2023prompt2prompt,tumanyan2023plugandplay}, and StyleID~\cite{chung2024styleid} demonstrated this most effectively.
It substitutes the key and value features of the content image's self-attention with those from a style image, then blends the queries with a mixing parameter $\gamma \in [0,1]$.

The limitation is straightforward: StyleID applies $\gamma$ uniformly across every decoder layer and every denoising timestep.
But these stages play different roles.
Shallow decoder layers encode coarse spatial layout while deeper layers govern fine-grained texture~\cite{tumanyan2023plugandplay}.
Early denoising steps determine global color and composition while later steps refine edges and structural detail~\cite{hertz2024editfriendly}.
A single fixed $\gamma$ treats all of this identically, locking the practitioner into one tradeoff point when the model is actually capable of much finer control.

Several recent works have begun to explore pieces of this.
AttenST~\cite{jiang2025attenst} shows that restricting style injection to specific decoder blocks already improves results, validating the idea that different layers matter differently.
StyleSSP~\cite{xu2025stylessp} and InstantStyle-Plus~\cite{he2024instantstyleplus} both combine ControlNet~\cite{zhang2023controlnet} conditioning with style injection to improve content preservation.
These are genuine contributions, but each proposes a complete pipeline and probes only one dimension.
None asks the more fundamental question: across all available axes of control - layers, timesteps, and geometric conditioning - are these effects independent, and can they be combined to reach tradeoffs no single technique achieves on its own?

We answer this through a systematic empirical study built on StyleID~\cite{chung2024styleid} as a controlled baseline.
Rather than proposing a new architecture, we treat StyleID's global $\gamma$ as one point in a richer design space and vary it across decoder layers, across denoising timesteps~\cite{song2021ddim}, and supplement it with ControlNet~\cite{zhang2023controlnet} geometric conditioning scheduled along both dimensions.
We begin with linear schedules to isolate the effect of dimension and direction from schedule shape, then extend to non-linear forms (cosine, square-root, quadratic, exponential) to further improve performance.

\noindent\textbf{Contributions.} We make the following contributions:
\begin{itemize}
    \item \textbf{Scheduled style injection.} A monotonically decreasing schedule on $\gamma$ across decoder layers or denoising timesteps consistently outperforms any fixed operating point, revealing a predictable gradient in style-content sensitivity across model depth and the denoising trajectory. Non-linear shapes (cosine, square-root) further improve over linear by concentrating the style-preservation transition in the most content-sensitive phase of denoising.
    \item \textbf{ControlNet conditioning scheduling.} Applying the same scheduling principle to ControlNet~\cite{zhang2023controlnet} conditioning scale provides a complementary content-preservation axis that operates independently of attention-based style injection.
    \item \textbf{Orthogonality and Pareto expansion.} Gamma scheduling and ControlNet conditioning compose nearly additively, confirming they act on independent dimensions. Their combination expands the achievable Pareto frontier beyond any single fixed-$\gamma$ baseline, with findings generalizing across SD\,1.4, 1.5, and 2.1 with identical rank ordering.
\end{itemize}

\Cref{sec:related} reviews related work. \Cref{sec:method} describes the method. \Cref{sec:experiments,sec:results} cover experimental setup and results. \Cref{sec:analysis} discusses mechanistic interpretations and \cref{sec:conclusion} concludes.

%% file: sec/2_related.tex
\section{Related Work}
\label{sec:related}

\subsection{Classical Neural Style Transfer}

Neural style transfer originates with Gatys \etal~\cite{gatys2016image}, who showed that Gram matrices of VGG~\cite{simonyan2015vgg} feature activations capture artistic style while the activations preserve spatial content; their subsequent analysis~\cite{gatys2017texture} established the relationship between deep texture synthesis and style perception.
Li \etal~\cite{li2017demystifying} proved that Gram matrix matching is equivalent to minimizing the Maximum Mean Discrepancy between feature distributions, establishing style as a fundamentally distributional property.
Huang and Belongie's AdaIN~\cite{huang2017adain} enabled arbitrary style transfer by matching channel-wise feature statistics in a single forward pass.
Attention-based methods including SANet~\cite{park2019sanet}, StyTR$^2$~\cite{chen2022stytr2}, AdaAttN~\cite{liu2021adaattn}, and ArtFlow~\cite{an2021artflow} introduced patch-level semantic matching and established the evaluation protocol (20 content $\times$ 40 style $= 800$ images, ArtFID~\cite{wright2022artfid}) that all subsequent work follows.
Geirhos \etal~\cite{geirhos2019texture} showed that deep networks are biased toward local texture over global shape, helping explain why key-value features in pretrained models are effective style carriers.

\subsection{Diffusion-Based Style Transfer}

Large-scale latent diffusion models~\cite{rombach2022ldm,podell2024sdxl} introduced a fundamentally different visual prior for style transfer.
Prompt-to-Prompt~\cite{hertz2023prompt2prompt} showed that cross-attention maps control spatial layout, and Plug-and-Play~\cite{tumanyan2023plugandplay} extended this to self-attention, demonstrating that spatial features carry structural information transferable across generation runs.
These works established a key intuition: what the model attends to determines what it generates, and this can be exploited without retraining.

Early diffusion-based style transfer still required optimization.
InST~\cite{zhang2023inst} combined DDIM inversion~\cite{song2021ddim} with textual inversion to encode style as learned text tokens, requiring several minutes per style.
DiffStyle~\cite{kwon2023diffstyle} attempted training-free transfer through h-space and skip connection manipulation, but found that naive feature injection causes content collapse.

StyleID~\cite{chung2024styleid} addressed this with training-free key-value substitution in self-attention, controlled by a query-blending parameter $\gamma$, achieving state-of-the-art ArtFID~\cite{wright2022artfid} and becoming the foundation concurrent work builds on. The limitation our work targets is that $\gamma$ is applied uniformly across all decoder layers and timesteps.

\subsection{Concurrent Training-Free Methods}

Several recent works have explored different angles on training-free style transfer.
AttenST~\cite{jiang2025attenst} found that restricting style injection to specific SDXL~\cite{podell2024sdxl} upsampling blocks already improves results (ArtFID 28.69 vs.\ StyleID's 28.80), confirming that different decoder layers play different roles; our layer-wise gamma scheduling extends this from a discrete block-selection choice to a continuous strength modulation.
DiffuseST~\cite{wang2024diffusest} introduces timestep-wise residual style features; our timestep scheduling is complementary, modulating StyleID's injection strength per timestep without architectural changes.
StyleSSP~\cite{xu2025stylessp} manipulates the frequency spectrum of the DDIM~\cite{song2021ddim} inversion latent to prevent content leakage - changing where denoising starts rather than what happens during it - making it orthogonal to our approach.
InstantStyle~\cite{wang2024instantstyle} and InstantStyle-Plus~\cite{he2024instantstyleplus} combine ControlNet~\cite{zhang2023controlnet} conditioning with style injection via IP-Adapter~\cite{ye2023ipadapter}, but require trainable components.
StyleAligned~\cite{jeong2024stylealigned} targets batch-level style consistency rather than single-pair transfer, though it confirms that self-attention key-value mechanisms are powerful style carriers.

Each of these works probes one axis of the style-content tradeoff.
None asks whether these axes are orthogonal or what happens when they are combined.

\subsection{Structural Conditioning}

ControlNet~\cite{zhang2023controlnet} adds a trainable encoder branch conditioned on structural signals like depth maps~\cite{Ranftl2022} or Canny edges, with T2I-Adapter~\cite{mou2024t2iadapter} as a lighter-weight variant; the core idea is that geometry and appearance can be conditioned separately.
In style transfer, ControlNet has been used for content preservation in StyleSSP~\cite{xu2025stylessp} and InstantStyle-Plus~\cite{he2024instantstyleplus}.
We use it similarly but with an analytic goal: by measuring how ControlNet affects content metrics independently of style metrics, we quantify how independently the two aspects are controllable.
Our results confirm they largely are (ControlNet improves LPIPS~\cite{zhang2018lpips} with minimal effect on FID~\cite{heusel2017fid}), making it a clean additional axis for Pareto frontier expansion.

\subsection{Industrial Applications}

Beyond artistic transfer, style transfer has become a practical tool for domain adaptation across industries.
CycleGAN~\cite{zhu2017unpaired} established unpaired image translation as a mechanism for sim-to-real transfer, with impact in autonomous driving~\cite{atapour2018real,arruda2022cross} (adapting synthetic or daytime data to real-world conditions) and medical imaging~\cite{iacono2023structure} (cross-modality MRI-to-CT translation without paired acquisitions).
More recently, Chigot \etal~\cite{chigot2025style} apply diffusion-based class-aware style transfer for synthetic-to-real semantic segmentation, showing robustness under adverse weather and lighting.
These applications share a common need for controllable appearance transfer, which is precisely what our scheduling framework aims to improve.

%% file: sec/3_method.tex
\section{Method}
\label{sec:method}

Our method builds directly on StyleID~\cite{chung2024styleid}, without changing its architecture or adding trainable parameters. We take its single global control knob ($\gamma$) and allow it to vary across decoder layers and denoising timesteps, and add ControlNet~\cite{zhang2023controlnet} as an independent content-preservation mechanism whose conditioning scale varies along the same two dimensions. This gives four scheduling axes in total, each described below following a brief recap of StyleID.

\subsection{Preliminaries: How StyleID Works}

StyleID~\cite{chung2024styleid} transfers artistic style by manipulating self-attention~\cite{vaswani2017attention} features inside a pre-trained Stable Diffusion~\cite{rombach2022ldm} model at inference time, with no training or optimization required. In the decoder's self-attention layers, query features ($Q$) encode spatial structure while key ($K$) and value ($V$) features encode texture and appearance; replacing the content's $K$ and $V$ with those from the style image produces an output that preserves spatial layout while adopting the style's texture.

Both content and style images are inverted into noise space via DDIM inversion~\cite{song2021ddim}, collecting query features $Q^c_t$ from the content and key-value features $K^s_t$, $V^s_t$ from the style at every timestep $t$. During the reverse denoising process, StyleID replaces the content's $K$ and $V$ with the style's in the self-attention layers of decoder layers 6 through 11.

Simply swapping $K$ and $V$ would destroy the content's structure over the course of denoising. To prevent this, StyleID introduces query preservation: it blends the content's original query with the stylized query using a mixing parameter $\gamma$:
\begin{equation}
    \tilde{Q}^{cs}_t = \gamma \cdot Q^c_t + (1 - \gamma) \cdot Q^{cs}_t
    \label{eq:query_preservation}
\end{equation}
The blended query is then used in the attention computation with the style's $K$ and $V$:
\begin{equation}
    \phi^{cs}_{out} = \text{Attn}(\tilde{Q}^{cs}_t,\, K^s_t,\, V^s_t)
    \label{eq:attn}
\end{equation}
Here, $\gamma$ controls the style-content balance. A higher $\gamma$ (closer to 1) preserves more content; a lower $\gamma$ (closer to 0) transfers more style. StyleID uses $\gamma = 0.75$ as the default and applies the same value at every decoder layer and every timestep.

StyleID includes two additional components: attention temperature scaling ($\tau = 1.5$), which compensates for reduced query-key similarity across images, and initial latent AdaIN~\cite{huang2017adain}, which normalizes the content noise to match the style's channel-wise statistics. Both are kept unchanged in all our experiments.

\subsection{Gamma Scheduling}

StyleID applies a fixed $\gamma$ uniformly across all decoder layers and timesteps. We relax this by letting $\gamma$ vary along two dimensions, decoder layers and denoising timesteps, using a unified scheduling framework.

All schedules interpolate $\gamma$ from a start value to an end value over $N$ positions. Let $\alpha_i = i/(N-1)$ for $i = 0, \dots, N-1$ denote the normalized position. A warping function $f\colon[0,1]\to[0,1]$ with $f(0)=0$, $f(1)=1$ gives:
\begin{equation}
    \gamma_i = \gamma_{\text{start}} + (\gamma_{\text{end}} - \gamma_{\text{start}}) \cdot f(\alpha_i)
    \label{eq:general_schedule}
\end{equation}
Decreasing schedules set $(\gamma_{\text{start}}, \gamma_{\text{end}}) = (\gamma_{\text{base}},\, \gamma_{\text{base}}/2)$, \eg $0.75 \to 0.375$; increasing schedules reverse this. Both directions are tested to confirm that direction matters, not merely the fact that $\gamma$ varies. We evaluate five warping functions:
\begin{equation}
    f(\alpha) \;\in\; \left\{\, \alpha,\;\; \alpha^{2},\;\; \sqrt{\alpha},\;\; \tfrac{1-\cos(\alpha\pi)}{2},\;\; \tfrac{e^{\alpha}-1}{e-1} \,\right\}
    \label{eq:schedule_shapes}
\end{equation}
The five functions appear in \cref{eq:schedule_shapes} in the order: linear, quadratic, square-root, cosine, exponential. They share the same endpoints but differ in where the transition is concentrated: quadratic is slow early and fast late; square-root is fast early and slow late; cosine is S-shaped with gradual transitions at both ends; exponential accelerates strongly toward the end. Linear serves as the controlled baseline when isolating the effect of scheduling dimension from schedule shape.

\subsubsection{Layer-wise Gamma Scheduling}
\label{sec:layer_gamma}

Layer-wise scheduling sets $N=6$ over decoder layers 6--11. Shallow layers encode coarse spatial structure while deeper layers handle fine-grained texture~\cite{tumanyan2023plugandplay}, motivating higher $\gamma$ at shallow layers, i.e., a decreasing schedule.

\subsubsection{Timestep-wise Gamma Scheduling}
\label{sec:timestep_gamma}

Timestep-wise scheduling sets $N=50$ over the denoising timesteps. Early high-noise timesteps establish global layout; later timesteps refine texture~\cite{hertz2024editfriendly}, motivating the same decreasing direction. The finer resolution of 50 timesteps over 6 layers allows closer alignment with the model's sensitivity curve, consistent with the stronger gains we observe for timestep scheduling in \cref{sec:results}.

\subsection{ControlNet Integration}

Gamma scheduling modulates how much style gets injected through the attention mechanism. A complementary way to preserve content is to provide the model with explicit geometric information about the content image, which is what ControlNet~\cite{zhang2023controlnet} does.

ControlNet adds a parallel encoder branch to the diffusion model, conditioned on a structural signal like a depth map~\cite{Ranftl2022} or Canny edge map. This branch feeds geometric information into the decoder at each step, guiding the model to maintain the content's spatial layout regardless of what the attention layers are doing with style. Depth maps are estimated from content images using MiDaS~\cite{Ranftl2022}.

The key hypothesis is that ControlNet and gamma scheduling address different aspects of the style-content tradeoff: gamma scheduling controls \emph{what} the model attends to (texture identity), while ControlNet controls \emph{where} things are placed (spatial layout). If these act on independent dimensions, their effects should compose without interfering with each other.

To test this, we explore ControlNet along two axes. First, we evaluate fixed conditioning scales of 0.25, 0.5, and 1.0 applied across decoder layers 6--11. Second, we apply the same scheduling framework of \cref{eq:general_schedule,eq:schedule_shapes} to the conditioning scale, treating it as a schedulable quantity in place of $\gamma$. Fixed scale 0.25 emerges as the most effective operating point; scheduling variants yield only minor improvements on content fidelity relative to their fixed baseline. Comprehensive results across all ControlNet configurations are provided in the supplementary material; the main paper reports the configurations that best complement gamma scheduling on the Pareto frontier.

%% file: sec/4_experiments.tex
\section{Experimental Setup}
\label{sec:experiments}

\subsection{Dataset and Evaluation Protocol}

We follow the evaluation protocol of StyleID~\cite{chung2024styleid} and StyTR$^2$~\cite{chen2022stytr2}, using 20 content images from MS-COCO~\cite{lin2014coco} and 40 style images from WikiArt~\cite{wikiart}, yielding 800 stylized outputs per experimental configuration. This shared protocol allows direct comparison with reported baselines without re-running them. All images are generated at $512 \times 512$ resolution.

\subsection{Evaluation Metrics}

We report four complementary metrics: ArtFID~\cite{wright2022artfid} (our primary combined metric, validated against human judgment with Spearman $\rho > 0.93$), FID~\cite{heusel2017fid} (style fidelity), LPIPS~\cite{zhang2018lpips} (content preservation), and CFSD~\cite{chung2024styleid} (structural content fidelity via VGG19 patch correlations). Together, ArtFID summarizes overall quality while FID captures the style axis and LPIPS with CFSD capture the content axis, letting us diagnose tradeoffs clearly across configurations.

\subsection{Implementation Details}

All experiments performed to demonstrate the expanding of the pareto frontier use Stable Diffusion v1.4~\cite{rombach2022ldm} as the backbone with DDIM sampling~\cite{song2021ddim} at 50 steps. No training or fine-tuning is performed at any stage. Style injection follows StyleID~\cite{chung2024styleid}: keys and values from the style image's DDIM inversion are injected into decoder layers 6 through 11. ControlNet experiments use ControlNet v1.0~\cite{zhang2023controlnet} with depth conditioning on decoder layers 6-11, where depth maps are estimated by MiDaS~\cite{Ranftl2022}. All experiments were run on a single NVIDIA RTX 3090. In total, we evaluated over 35 configurations covering the baseline, individual scheduling axes, and their combinations, producing more than 28,000 stylized images.

%% file: sec/5_results.tex
\section{Results}
\label{sec:results}

\begin{figure*}[t]
\centering
\includegraphics[width=\linewidth]{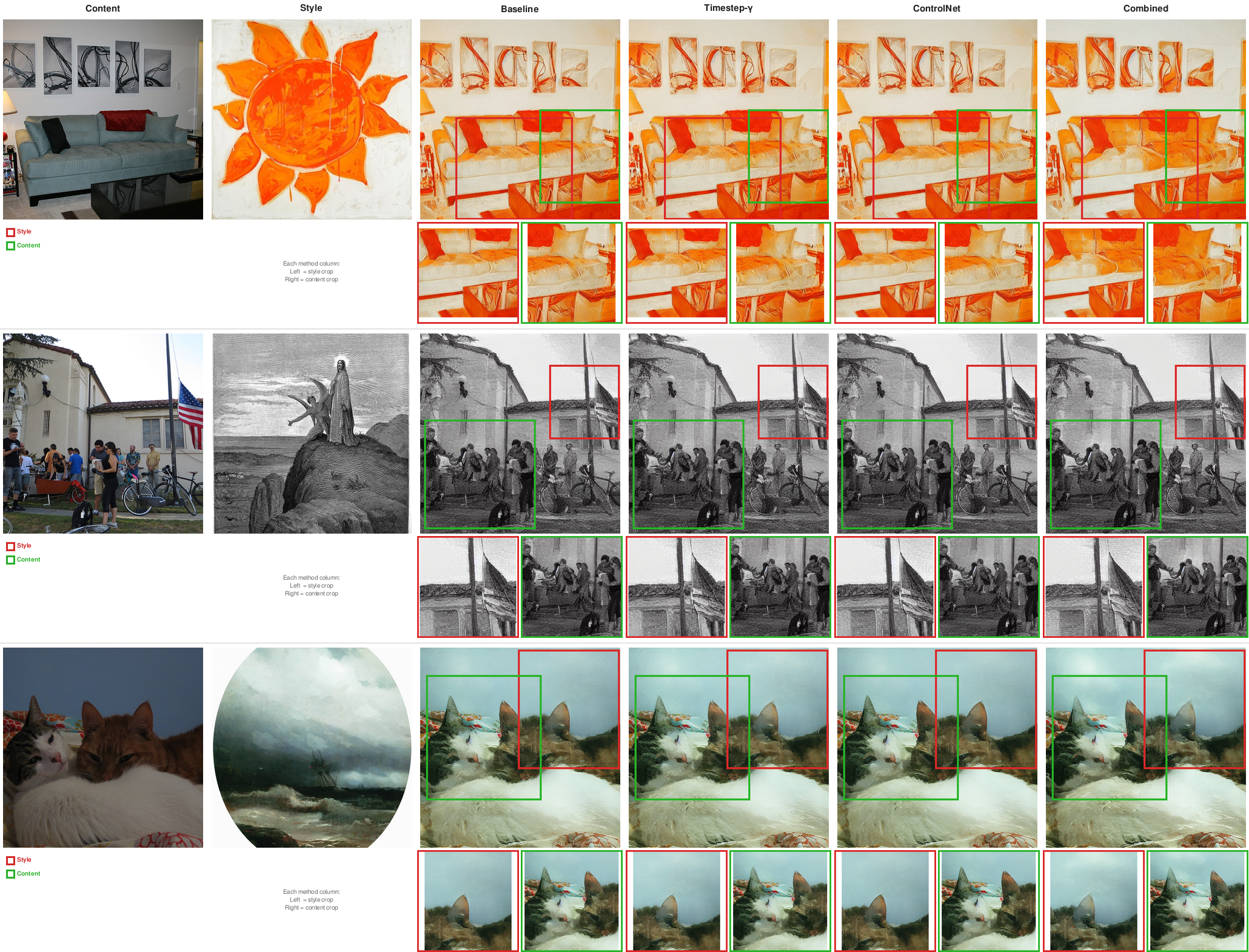}
\caption{Qualitative ablation across three content-style pairs.
\textbf{Baseline}: StyleID~\cite{chung2024styleid} with fixed $\gamma{=}0.75$.
\textbf{Timestep-$\gamma$}: best cosine schedule ($\gamma\!:0.75{\to}0.375$, top entry in \cref{tab:ablation}), improving style fidelity.
\textbf{ControlNet}: depth conditioning ($\text{CN}{=}0.25$), tightening content adherence.
\textbf{Combined}: our best configuration (Ours, \cref{tab:comparison}).
Per-method zoomed crops are shown below each row: \textit{red-outlined} highlights style-discriminative regions; \textit{green-outlined} highlights content-sensitive regions.
Combined achieves a good balance across all pairs; individual component gains are subtle but visible in the crops.}
\label{fig:qualitative}
\end{figure*}

We report results across four axes: a quantitative comparison against prior methods (\cref{tab:comparison}), a qualitative comparison of our scheduling strategies (\cref{fig:qualitative}), an ablation over individual scheduling axes (\cref{tab:ablation}), and a backbone generalization study across SD\,1.4, 1.5, and 2.1 (\cref{tab:backbone}).
All core results use $\gamma_\text{base}{=}0.75$ to match StyleID's default operating point and enable direct comparison against reported baselines.
A full sweep across all gamma values (0.50, 0.75, 0.90) and all configurations is provided in the supplementary material.

\noindent\textit{Quantitative comparison.}
\cref{tab:comparison} compares ArtFID, FID, LPIPS, and CFSD across 12 prior methods and our two best configurations on 800 content-style pairs.
Both outperform all compared methods on ArtFID.
The cosine timestep-$\gamma$ schedule alone achieves the best overall ArtFID of 26.976 and the lowest FID of 16.124 among all evaluated methods, a 6.3\% ArtFID improvement over the StyleID baseline (28.801).
The combined configuration ($\sqrt{\cdot}$-$\gamma$ with a timestep-scheduled ControlNet depth signal) reaches ArtFID 27.036 - a marginal ArtFID trade-off relative to $\gamma$-only - while recovering content fidelity (LPIPS: 0.575 $\to$ 0.564; CFSD: 0.297 $\to$ 0.295), confirming that the two mechanisms operate on complementary axes.

\begin{table*}[t]
\centering
\caption{%
Quantitative comparison on 800 images. Cols.\ 2--9: conventional; cols.\ 10--13: diffusion-based; col.~14: ours (cos-$\gamma$, TS$\downarrow$); col.~15: ours ($\sqrt{\cdot}$-$\gamma$ + CN, both TS$\downarrow$; CN: $0.25\!\to\!0.125$). Baseline from~\cite{chung2024styleid}; \textbf{bold} = best; \underline{underline} = second-best.%
}
\label{tab:comparison}
\setlength{\tabcolsep}{3pt}
\renewcommand{\arraystretch}{1.05}
\footnotesize
\begin{tabular}{@{}l
    c c c c c c c c
    c c c c
    c c
  @{}}
\toprule
\textbf{Metric}
  & \makecell{AdaIN\\\cite{huang2017adain}}
  & \makecell{ArtFlow\\\cite{an2021artflow}}
  & \makecell{CAST\\\cite{kim2022cast}}
  & \makecell{EFDM\\\cite{zhang2022efdm}}
  & \makecell{MAST\\\cite{deng2020mast}}
  & \makecell{AdaAttN\\\cite{liu2021adaattn}}
  & \makecell{StyTR$^2$\\\cite{chen2022stytr2}}
  & \makecell{AesPA-Net\\\cite{hong2023aespanet}}
  & \makecell{InST\\\cite{zhang2023inst}}
  & \makecell{DiffStyle\\\cite{kwon2023diffstyle}}
  & \makecell{StyleID\\\cite{chung2024styleid}}
  & \makecell{AttenST\\\cite{jiang2025attenst}}
  & \makecell{Ours\\(cos-$\gamma$)}
  & \makecell{Ours\\($\sqrt{\cdot}$-$\gamma$+CN$\downarrow$)} \\
\midrule
ArtFID $\downarrow$
  & 30.933 & 34.630 & 34.685 & 34.605 & 31.282 & 30.350 & 30.720 & 31.420
  & 40.633 & 41.464 & 28.801 & 28.693
  & \textbf{26.976} & \underline{27.036} \\
FID $\downarrow$
  & 18.242 & 21.252 & 20.395 & 20.062 & 18.199 & 18.658 & 18.890 & 19.760
  & 21.571 & 20.903 & 18.131 & 18.559
  & \textbf{16.124} & \underline{16.285} \\
LPIPS $\downarrow$
  & 0.608  & 0.556  & 0.621  & 0.643  & 0.629  & 0.544  & 0.545  & 0.514
  & 0.800  & 0.893  & \underline{0.505} & \textbf{0.467}
  & 0.575  & 0.564 \\
CFSD $\downarrow$
  & 0.315    & 0.292    & 0.292    & 0.335    & 0.304    & 0.286    & 0.301    & \underline{0.246}
  & 0.676    & 0.282    & \textbf{0.228} & ---
  & 0.297  & 0.295 \\
\bottomrule
\end{tabular}
\end{table*}

\begin{figure}[t]
\centering
\includegraphics[width=\linewidth]{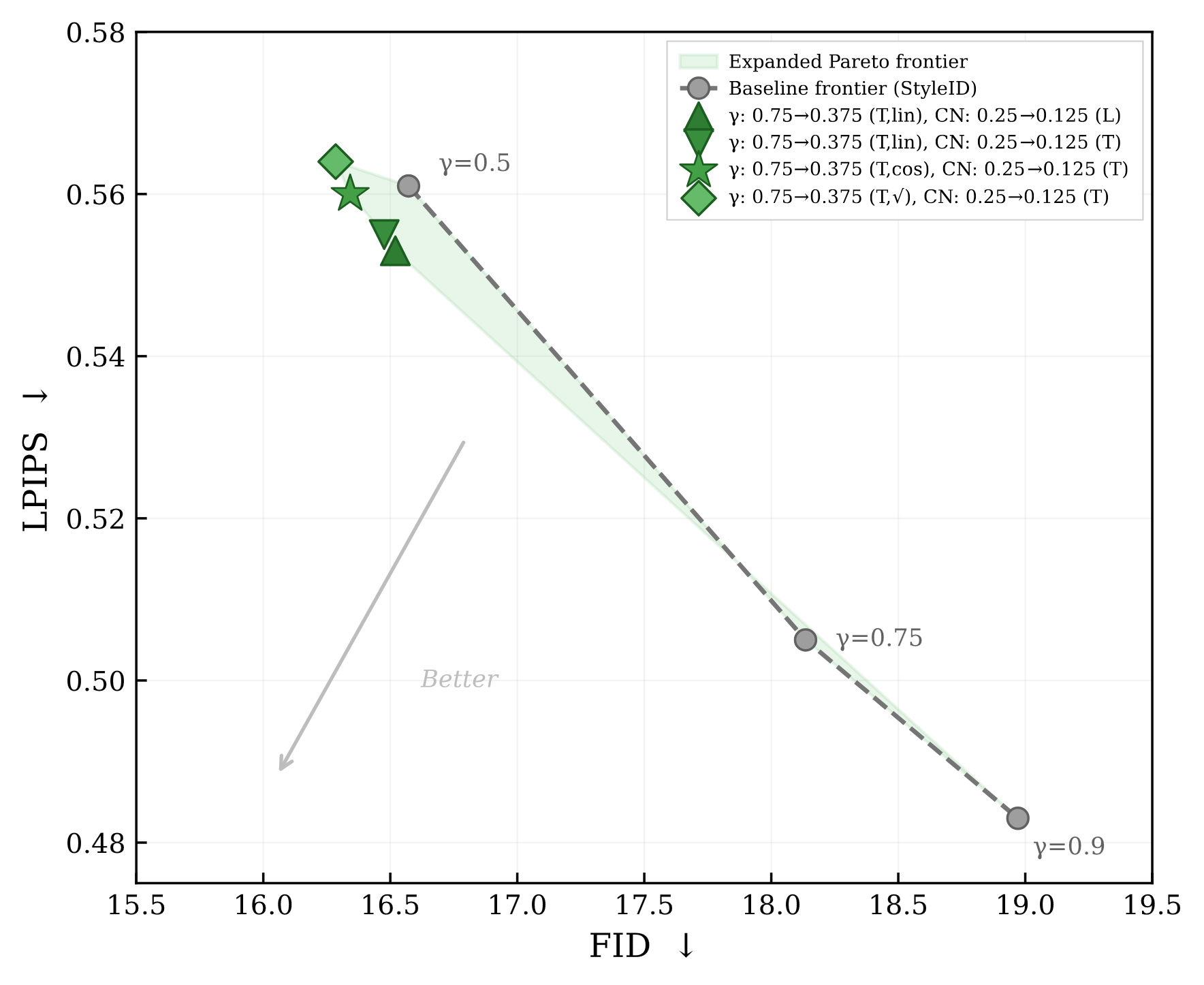}
\caption{Pareto frontier comparison between baseline StyleID and proposed combined scheduling methods. The gray dotted line represents the baseline frontier formed by StyleID with different gamma values ($\gamma$=0.5, 0.75, 0.9). The green symbols show our combined experiments that expand the Pareto frontier. The light green shaded region highlights the expanded frontier area}
\label{fig:pareto}
\end{figure}

\noindent\textit{Pareto expansion.}
\cref{fig:pareto} visualizes this as a Pareto frontier expansion in FID--LPIPS space.
The baseline StyleID frontier (gray) represents the best achievable trade-offs under fixed uniform gamma.
Our combined configurations (green) lie strictly outside this frontier, demonstrating that our method expands the frontier rather than merely shifting the operating point along it.

\noindent\textit{Qualitative comparison.}
\cref{fig:qualitative} provides a qualitative view of the same progression.
Timestep-$\gamma$ scheduling produces richer stylization relative to the baseline, while ControlNet depth conditioning tightens spatial coherence.
The combined method achieves the most balanced result across all content-style pairs, with improvements most visible in fine texture and structural boundaries.

\noindent\textit{Backbone generalization.}
To assess whether the scheduling effects are specific to SD\,1.4, we evaluate the core configurations on SD\,1.5 and SD\,2.1 (\cref{tab:backbone}).
The rank ordering of all configurations is identical across all three backbones, with decreasing timestep schedules yielding 5.0--6.0\% ArtFID gains and the combined method achieving 4.6--5.7\% improvement across the family, confirming that the findings reflect the shared decoder structure of the Stable Diffusion family rather than SD\,1.4-specific behaviour.

\noindent\textit{Scope of comparison.}
Several concurrent methods differ from ours in base architecture (e.g., SDXL vs.\ SD\,1.4), use trainable components, or introduce architectural modifications beyond scheduling.
Direct numerical comparison across these differences would conflate scheduling gains with architectural ones; our evaluation instead isolates the scheduling dimension within a consistent, purely training-free framework built on StyleID.

\input{ablation_table_gamma75}

%% file: ablation_table_gamma75.tex
\begin{table*}[t]
\centering
\caption{%
Ablation at $\gamma_\text{base}{=}0.75$, four groups. \textbf{G1}: gamma direction - L (layerwise) and T (timestep), $\downarrow$ / $\uparrow$ = decreasing/increasing, all linear. \textbf{G2}: schedule shape on timestep axis, range $0.75\!\to\!0.375$; $\dagger$ = linear from G1 repeated for reference. \textbf{G3}: ControlNet depth in isolation. \textbf{G4}: combined gamma + CN; (cos)/($\sqrt{\cdot}$) = cosine/sqrt shapes. $a \xrightarrow{\text{L}} b$: layerwise (layers 6--11); $a \xrightarrow{\text{T}} b$: timestep (50 steps). \textbf{Bold} = best per column.%
}
\label{tab:ablation}
\small
\setlength{\tabcolsep}{5pt}
\begin{tabular}{@{}llcccc@{}}
\toprule
\textbf{Axis} & \textbf{Schedule} & \textbf{ArtFID}~$\downarrow$ & \textbf{FID}~$\downarrow$ & \textbf{LPIPS}~$\downarrow$ & \textbf{CFSD}~$\downarrow$ \\
\midrule
Baseline & $\gamma = 0.75$ (fixed) & 28.806 & 18.135 & 0.505 & 0.228 \\
Baseline & $\gamma = 0.50$ (fixed) & 27.430 & 16.572 & 0.561 & 0.279 \\
\midrule
Gamma (L) $\downarrow$ & $0.75 \xrightarrow{\text{L}} 0.375$ & 27.471 & 16.506 & 0.569 & 0.287 \\
Gamma (L) $\uparrow$   & $0.375 \xrightarrow{\text{L}} 0.75$  & 28.249 & 17.546 & 0.523 & 0.244 \\
Gamma (T) $\downarrow$ & $0.75 \xrightarrow{\text{T}} 0.375$  & 27.089 & 16.250 & 0.570 & 0.291 \\
Gamma (T) $\uparrow$   & $0.375 \xrightarrow{\text{T}} 0.75$  & 28.530 & 17.735 & 0.523 & 0.243 \\
\midrule
Gamma (T) $\downarrow$ & $0.75 \xrightarrow{\text{T}} 0.375$~(lin)$^{\dagger}$  & 27.089 & 16.250 & 0.570 & 0.291 \\
Gamma (T) $\downarrow$ & $0.75 \xrightarrow{\text{T}} 0.375$~(exp)              & 27.182 & 16.389 & 0.563 & 0.283 \\
Gamma (T) $\downarrow$ & $0.75 \xrightarrow{\text{T}} 0.375$~(quad)             & 27.270 & 16.515 & 0.557 & 0.276 \\
Gamma (T) $\downarrow$ & $0.75 \xrightarrow{\text{T}} 0.375$~(cos)              & \textbf{26.976} & 16.124 & 0.575 & 0.297 \\
Gamma (T) $\downarrow$ & $0.75 \xrightarrow{\text{T}} 0.375$~($\sqrt{\cdot}$)   & 26.995 & \textbf{16.067} & 0.582 & 0.304 \\
\midrule
CN (fixed)             & $\text{CN} = 0.25$                                      & 29.111 & 18.460 & \textbf{0.496} & 0.225 \\
CN (L) $\downarrow$    & $0.25 \xrightarrow{\text{L}} 0.125$                     & 29.054 & 18.424 & \textbf{0.496} & \textbf{0.224} \\
CN (T) $\downarrow$    & $0.25 \xrightarrow{\text{T}} 0.125$                     & 29.032 & 18.384 & 0.498 & 0.225 \\
\midrule
Combined & $\gamma\!: 0.75 \xrightarrow{\text{T}} 0.375$~(lin),~$\text{CN}\!: 0.25 \xrightarrow{\text{L}} 0.125$                   & 27.207 & 16.520 & 0.553 & 0.282 \\
Combined & $\gamma\!: 0.75 \xrightarrow{\text{T}} 0.375$~(lin),~$\text{CN}\!: 0.25 \xrightarrow{\text{T}} 0.125$                   & 27.183 & 16.476 & 0.555 & 0.284 \\
Combined & $\gamma\!: 0.75 \xrightarrow{\text{T}} 0.375$~(cos),~$\text{CN}\!: 0.25 \xrightarrow{\text{T}} 0.125$             & 27.052 & 16.342 & 0.560 & 0.290 \\
Combined & $\gamma\!: 0.75 \xrightarrow{\text{T}} 0.375$~($\sqrt{\cdot}$),~$\text{CN}\!: 0.25 \xrightarrow{\text{T}} 0.125$ & 27.036 & 16.285 & 0.564 & 0.295 \\
\bottomrule
\end{tabular}
\end{table*}

\begin{table*}[t]
\centering
\small
\setlength{\tabcolsep}{3.5pt}
\caption{Scheduling generalization across SD backbones ($\gamma_\text{base}{=}0.75$).
$\Delta\%$ in parentheses = ArtFID change relative to each backbone's own baseline.
The rank ordering of all configurations is identical across all three backbones.
CN fixed uses ControlNet v1.0~\cite{zhang2023controlnet} for SD\,1.4, v1.1~\cite{zhang2023controlnet} for SD\,1.5, and thibaud/controlnet-sd21~\cite{thibaud2023controlnetsd21} for SD\,2.1.}
\label{tab:backbone}
\begin{tabular}{@{}lccccccccc@{}}
\toprule
 & \multicolumn{3}{c}{\textbf{SD\,1.4}}
 & \multicolumn{3}{c}{\textbf{SD\,1.5}}
 & \multicolumn{3}{c}{\textbf{SD\,2.1}} \\
\cmidrule(lr){2-4}\cmidrule(lr){5-7}\cmidrule(lr){8-10}
\textbf{Config}
  & \makecell{ArtFID$\downarrow$} & FID$\downarrow$ & \makecell{LPIPS$\downarrow$}
  & \makecell{ArtFID$\downarrow$} & FID$\downarrow$ & \makecell{LPIPS$\downarrow$}
  & \makecell{ArtFID$\downarrow$} & FID$\downarrow$ & \makecell{LPIPS$\downarrow$} \\
\midrule
Baseline ($\gamma{=}0.75$)
  & 28.81 & 18.13 & 0.505
  & 28.79 & 18.10 & 0.508
  & 29.89 & 19.69 & 0.445 \\
\midrule
$\gamma$: $0.75 \xrightarrow{\text{L}} 0.375$
  & 27.47\gain{4.7} & 16.51 & 0.569
  & 27.49\gain{4.5} & 16.49 & 0.572
  & 28.66\gain{4.1} & 18.35 & 0.481 \\
$\gamma$: $0.375 \xrightarrow{\text{L}} 0.75$
  & 28.25\gain{2.0} & 17.55 & 0.523
  & 28.23\gain{2.0} & 17.51 & 0.525
  & 29.40\gain{1.6} & 19.21 & 0.455 \\
$\gamma$: $0.75 \xrightarrow{\text{T}} 0.375$
  & 27.09\gain{6.0} & 16.25 & 0.570
  & 27.07\gain{6.0} & 16.22 & 0.572
  & 28.38\gain{5.0} & 18.08 & 0.487 \\
$\gamma$: $0.375 \xrightarrow{\text{T}} 0.75$
  & 28.53\gain{1.0} & 17.73 & 0.523
  & 28.53\gain{0.9} & 17.71 & 0.525
  & 29.65\gain{0.8} & 19.44 & 0.450 \\
CN fixed ($s{=}0.25$)
  & 29.11\loss{1.0} & 18.46 & 0.496
  & 29.07\loss{1.0} & 18.40 & 0.498
  & 30.02\loss{0.4} & 19.81 & 0.443 \\
\bottomrule
\end{tabular}
\end{table*}

%% file: sec/6_analysis.tex
\section{Analysis and Discussion}
\label{sec:analysis}

\subsection{Scheduling Direction Reflects the Diffusion U-Net's Sensitivity Hierarchy}

Decreasing gamma schedules outperform increasing ones in every tested configuration, across both decoder layers and denoising timesteps, without exception (see \cref{tab:ablation}).
This consistency is not incidental; it reflects a known structural property of diffusion U-Nets.
Tumanyan \etal~\cite{tumanyan2023plugandplay} demonstrated that early-to-middle decoder blocks encode coarse spatial structure, while later blocks encode local texture.
Prompt-to-Prompt~\cite{hertz2023prompt2prompt} established that attention maps in these later blocks are the primary carriers of fine-grained appearance.
Together, these findings suggest that early layers and early timesteps are content-sensitive, while later stages are style-tolerant.

A decreasing gamma schedule respects this hierarchy: query preservation is strongest in the stages that set coarse spatial structure and relaxes in the stages already committed to texture, as also visible in \cref{fig:qualitative}.
The reverse schedule weakens content protection precisely where the structural scaffold is being established, producing a consistent performance penalty across all base gamma values and both scheduling dimensions.
Timestep scheduling yields a larger margin than layerwise scheduling, consistent with its higher resolution: 50 timesteps allow closer alignment with the model's sensitivity curve than 6 decoder layers.
Consistent with the finding that deep visual networks are strongly biased toward texture over shape~\cite{geirhos2019texture}, our results add a scheduling corollary: \emph{the strength of style injection should be conditioned on where and when in the model it is applied.}

A natural question is whether the gains from gamma scheduling simply reflect the effect of using a lower average gamma, rather than the schedule itself.
The data argue against this interpretation.
The timestep schedule $0.75 \to 0.375$ (linear) has a mean gamma of approximately 0.5625, yet it achieves ArtFID 27.089 compared to 27.430 for the fixed $\gamma = 0.50$ baseline, which operates at a strictly lower average gamma.
The schedule achieves stronger style transfer despite the higher average gamma value, while remaining competitive on content preservation (LPIPS 0.570 vs.\ 0.561).
The layerwise schedule shows the same pattern: mean gamma ${\approx}0.5625$, ArtFID 27.471 versus 27.430 for fixed $\gamma = 0.50$, with better content preservation (LPIPS 0.569 vs.\ 0.561).
A uniform reduction in gamma cannot reproduce this behavior.

The same direction sensitivity extends to the schedule shape itself.
Among the five functional forms evaluated on the timestep axis (exponential, linear, quadratic, cosine, square-root), cosine and square-root achieve the best ArtFID of 26.976 and 26.995 respectively, compared to 27.089 for linear (see \cref{tab:ablation}).
All five are decreasing schedules over the same range ($0.75 \to 0.375$); the differences arise entirely from how steeply gamma decreases in the earliest timesteps, when content structure is most vulnerable.
Cosine and square-root both apply a gentler initial decrease followed by a sharper drop, keeping query preservation high precisely when the structural scaffold is being laid down, then relaxing it more aggressively in the later texture-refining steps.

\subsection{Gamma Scheduling and ControlNet Conditioning Operate on Near-Orthogonal Axes}

The second major finding is that gamma scheduling and ControlNet conditioning improve the style-content tradeoff through genuinely independent mechanisms. When two interventions are orthogonal, their individual gains add: you get the full benefit of each without one undermining the other, and you can tune them separately rather than searching their joint space.

The independence has a clear mechanistic basis. Gamma modulates the blended query $\tilde{Q}^{cs}_t = \gamma Q^c_t + (1-\gamma)Q^{cs}_t$, changing texture identity through a purely distributional operation with no direct effect on spatial layout. ControlNet~\cite{zhang2023controlnet} injects a depth map upstream of attention, constraining spatial placement without influencing feature selection. Because they act on different properties at different points in the computational graph, their effects do not interfere.

We verify this empirically using the linear schedule as a controlled case. Timestep-$\gamma$ scheduling drives the dominant FID improvement ($18.135 \to 16.250$, a gain of 1.885 points) at the cost of modest content degradation (LPIPS: $0.505 \to 0.570$, a loss of 0.065). ControlNet conditioning alone shows the complementary profile: it recovers content fidelity (LPIPS: $0.505 \to 0.498$) with negligible style benefit. Under perfect additivity, the combined configuration should recover the LPIPS cost of gamma scheduling while preserving its FID gain. The actual combined result (FID $= 16.476$, LPIPS $= 0.555$) closely matches this prediction, with LPIPS partially recovering as expected while FID remains near the gamma-only result, confirming the two axes compose nearly additively.

Critically, this orthogonality holds across all non-linear schedule shapes. For the cosine schedule, gamma alone achieves FID $= 16.124$ with LPIPS $= 0.575$; adding CN scheduling recovers LPIPS to 0.560 while FID rises only slightly to 16.342. The square-root schedule behaves identically: gamma alone reaches FID $= 16.067$ with LPIPS $= 0.582$; the combined result is FID $= 16.285$ and LPIPS $= 0.564$. In every case, ControlNet acts as a content-recovery axis that does not meaningfully disturb the style gains established by gamma scheduling. The consistency across schedule shapes strengthens the orthogonality claim: the independence is not an artifact of the linear schedule but a structural property of how these two mechanisms interact.

This separation explains the Pareto expansion in \cref{fig:pareto}. No single fixed-$\gamma$ configuration can simultaneously achieve low FID and low LPIPS. The combined configurations escape this constraint: the square-root combined result achieves FID $= 16.285$ and LPIPS $= 0.564$, strictly better on both axes than the strongest single-axis baseline (fixed $\gamma = 0.50$: FID $= 16.572$, LPIPS $= 0.561$), and yields the best overall ArtFID of 27.036 across all evaluated configurations.

\subsection{Generalization Across Backbones}
\label{sec:backbone}

A natural question is whether the scheduling effects are specific to SD\,1.4 or reflect something more general about the Stable Diffusion family.
Table~\ref{tab:backbone} evaluates the core configurations on SD\,1.4, SD\,1.5, and SD\,2.1.
The $\Delta$\% values are computed relative to each backbone's own fixed baseline, which normalizes for the different absolute score scales across architectures.

Two findings hold without exception across all three backbones.
First, decreasing schedules consistently outperform increasing ones by a large margin: timestep-$\downarrow$ yields ArtFID gains of 5.0--6.0\%, while timestep-$\uparrow$ yields only 0.8--1.0\%.
The direction of the schedule matters far more than which backbone it is applied to.
Second, the rank ordering of all configurations is identical across SD\,1.4, SD\,1.5, and SD\,2.1.
CN fixed at $s{=}0.25$ shows a consistent modest ArtFID degradation of approximately $+$1.0\% for SD\,1.4 and SD\,1.5 and $+$0.4\% for SD\,2.1 relative to the baseline, while recovering LPIPS across all three backbones, confirming that the content-preservation behaviour of ControlNet conditioning generalizes across the Stable Diffusion family.

%% file: sec/7_conclusion.tex
\section{Conclusion}
\label{sec:conclusion}

We investigated whether StyleID's~\cite{chung2024styleid} fixed global gamma can be replaced by a schedule that varies injection strength across decoder layers or denoising timesteps. Both the direction and functional form of the schedule matter. Decreasing schedules outperform increasing ones in every tested configuration, consistent with the known early-content / late-style sensitivity gradient in diffusion U-Nets~\cite{tumanyan2023plugandplay,hertz2024editfriendly}. Among schedule shapes, cosine and square-root timestep schedules provide the best results by concentrating content protection in the most structure-sensitive early timesteps, achieving a 6.3\% ArtFID improvement over the StyleID baseline without any additional parameters.

ControlNet~\cite{zhang2023controlnet} depth conditioning operates on a near-orthogonal axis to gamma scheduling, recovering content structure that gamma scheduling trades away. Their improvements compose nearly additively, confirming that the two axes can be tuned independently. The best combined configuration expands the Pareto frontier to simultaneously lower FID and lower LPIPS than any single fixed-$\gamma$ baseline.

These findings generalize across SD\,1.4, 1.5, and 2.1 with identical rank ordering, confirming they reflect the shared decoder structure of the Stable Diffusion family. The broader implication is that the style-content tradeoff in training-free diffusion style transfer has at least two largely independent dimensions, and treating them separately expands the achievable Pareto frontier without modifying the underlying model. Future work includes validating on SDXL~\cite{podell2024sdxl} and conducting perceptual user studies to complement the quantitative evaluation.

%% file: sec/X_suppl.tex
\twocolumn[{%
  \begin{center}
    {\LARGE\textbf{Supplementary Material:}}\\[4pt]
    {\large\textbf{Scheduled Style Injection: Expanding the Style-Content Pareto Frontier\\
    in Training-Free Diffusion-based Style Transfer}}\\[12pt]
    Amey Sunil Kulkarni\\
    Independent Researcher\\
    {\tt\small amey1695@gmail.com}
  \end{center}
  \bigskip
}]

\setcounter{section}{0}
\renewcommand{\thesection}{S\arabic{section}}
\setcounter{table}{0}
\renewcommand{\thetable}{S\arabic{table}}

\noindent\textbf{Overview.}
The main paper argues that applying StyleID's style-injection parameter $\gamma$ uniformly
across all decoder layers and denoising timesteps is unnecessarily restrictive.
Because early denoising steps establish coarse spatial structure while later steps refine
texture, and because shallow decoder layers govern layout while deeper layers govern
appearance, a single fixed $\gamma$ treats structurally distinct computations identically.
Replacing it with a monotonically decreasing schedule --- stronger content preservation
early and in shallow layers, relaxing as the model transitions to texture --- consistently
improves the style-content tradeoff.
Adding ControlNet depth conditioning provides a complementary content-recovery axis that
acts on spatial layout independently of the attention-based style signal.
Because the two mechanisms operate on different properties at different points in the
computational graph, they compose nearly additively, expanding the achievable
Pareto frontier beyond what either achieves alone.

The main paper anchors all ablations at $\gamma_\text{base}{=}0.75$ to match StyleID's
published default, and selects depth conditioning at ControlNet scale~0.25 as the primary
CN configuration.
Both choices are deliberate but warrant independent validation: Is the scheduling advantage
specific to $\gamma{=}0.75$, or does it generalize? Does it grow or shrink at other
operating points? Is scale~0.25 genuinely optimal? Which conditioning type best preserves
style? Does scheduling the CN signal yield additional gains, and does schedule
\emph{shape} matter for CN as it does for gamma?

This supplementary answers each of these questions with the complete experimental
evidence. \textbf{S1} extends the gamma sweep to
$\gamma{=}0.50$ (style-heavy) and $\gamma{=}0.90$ (content-heavy), showing that
the directional advantage of decreasing schedules not only holds but grows with higher
baseline gamma, and identifying the global best single-mechanism result.
\textbf{S2} provides the full ControlNet variant grid: conditioning type, scale,
scheduling direction, and schedule shape --- validating the main paper's choices and
establishing that CN schedule shape and direction both have negligible effect, a null
result that confirms the orthogonality of the two improvement axes.

All evaluations use 800 content-style pairs (20 MS-COCO $\times$ 40 WikiArt images).
Metrics: ArtFID, FID, LPIPS, CFSD. All lower-is-better ($\downarrow$).
\textbf{Bold} = best ArtFID in each block.

\section{Full Gamma Sweep Across All Base Values}
\label{sec:suppl_gamma}

The main paper reports results only at $\gamma_\text{base}{=}0.75$.
The two tables below extend the evaluation to $\gamma{=}0.50$ (style-heavy) and
$\gamma{=}0.90$ (content-heavy), probing whether the scheduling advantage is specific to
the default operating point or reflects a deeper structural property of the model.
The answer is the latter: decreasing schedules consistently outperform increasing ones at
every base value, and the margin \emph{grows} with $\gamma$, which is exactly what the
sensitivity-hierarchy argument in the main paper predicts --- higher fixed gamma
means more aggressive uniform injection, so there is more structural damage to recover
by concentrating content protection early.

\paragraph{Layerwise scheduling (Table~\ref{tab:suppl_layer}).}
The layerwise schedule varies $\gamma$ linearly across decoder layers 6--11 (six steps).
Even with this coarse, six-point grid, the directional advantage is consistent: L$\downarrow$
beats L$\uparrow$ by 0.27 ArtFID at $\gamma{=}0.50$, 0.78 at $\gamma{=}0.75$, and
1.06 at $\gamma{=}0.90$.
The growing gap confirms that the sensitivity hierarchy is not a statistical fluke: it
scales predictably with how aggressively style is injected.
The smaller margin compared to the timestep axis (0.78 vs.\ 1.44 at $\gamma{=}0.75$)
reflects the coarser resolution of six layers versus 50 timesteps --- the schedule cannot
track the model's sensitivity curve as closely.

\begin{table}[t]
\centering
\footnotesize
\setlength{\tabcolsep}{3pt}
\caption{%
Layerwise gamma scheduling across all three base values.
Decreasing schedules outperform increasing ones at every base value;
the gap grows from 0.27 at $\gamma{=}0.50$ to 1.06 at $\gamma{=}0.90$.
\textbf{Bold} = best ArtFID per block.%
}
\label{tab:suppl_layer}
\begin{tabular}{@{}lrrrr@{}}
\toprule
\textbf{Schedule} & \textbf{ArtFID}$\downarrow$ & \textbf{FID}$\downarrow$ & \textbf{LPIPS}$\downarrow$ & \textbf{CFSD}$\downarrow$ \\
\midrule
Baseline $\gamma{=}0.50$              & 27.430 & 16.572 & 0.561 & 0.279 \\
L$\downarrow$ $0.50\!\to\!0.25$       & \textbf{26.739} & 15.500 & 0.621 & 0.355 \\
L$\uparrow$\phantom{$\downarrow$} $0.25\!\to\!0.50$ & 27.006 & 16.085 & 0.581 & 0.301 \\
\midrule
Baseline $\gamma{=}0.75$              & 28.806 & 18.135 & 0.505 & 0.228 \\
L$\downarrow$ $0.75\!\to\!0.375$      & \textbf{27.471} & 16.506 & 0.569 & 0.287 \\
L$\uparrow$\phantom{$\downarrow$} $0.375\!\to\!0.75$ & 28.249 & 17.546 & 0.523 & 0.244 \\
\midrule
Baseline $\gamma{=}0.90$              & 29.623 & 18.971 & 0.483 & 0.211 \\
L$\downarrow$ $0.90\!\to\!0.45$       & \textbf{27.925} & 17.078 & 0.545 & 0.262 \\
L$\uparrow$\phantom{$\downarrow$} $0.45\!\to\!0.90$ & 28.987 & 18.335 & 0.499 & 0.223 \\
\bottomrule
\end{tabular}
\end{table}

\paragraph{Timestep scheduling (Table~\ref{tab:suppl_ts}).}
The timestep schedule varies $\gamma$ linearly across all 50 DDIM steps, providing a
much finer grid than the six-layer counterpart.
The finer resolution translates directly into larger margins: the TS$\downarrow$ vs.\
TS$\uparrow$ gap is 0.77 at $\gamma{=}0.50$, widens to 1.44 at $\gamma{=}0.75$, and
reaches 1.79 at $\gamma{=}0.90$.
Mechanistically, the timestep axis captures the denoising trajectory's sensitivity curve
more faithfully: the earliest timesteps are the most content-critical (global composition
is fixed then), and TS$\downarrow$ maximally protects those steps while gradually relaxing
as the generation shifts to texture refinement.
The global best single-mechanism result across the entire experiment suite is
TS$\downarrow$ at $\gamma_\text{base}{=}0.50$: \textbf{ArtFID~26.459}.

\begin{table}[t]
\centering
\footnotesize
\setlength{\tabcolsep}{3pt}
\caption{%
Timestep gamma scheduling across all three base values.
TS$\downarrow$ at $\gamma{=}0.50$ achieves ArtFID 26.459 --- the best single-mechanism
result in the full experiment grid.
The $\downarrow$ vs $\uparrow$ gap grows from 0.77 to 1.44 to 1.79 as $\gamma$ increases
from 0.50 to 0.75 to 0.90.
\textbf{Bold} = best ArtFID per block.%
}
\label{tab:suppl_ts}
\begin{tabular}{@{}lrrrr@{}}
\toprule
\textbf{Schedule} & \textbf{ArtFID}$\downarrow$ & \textbf{FID}$\downarrow$ & \textbf{LPIPS}$\downarrow$ & \textbf{CFSD}$\downarrow$ \\
\midrule
Baseline $\gamma{=}0.50$               & 27.430 & 16.572 & 0.561 & 0.279 \\
TS$\downarrow$ $0.50\!\to\!0.25$       & \textbf{26.459} & 15.327 & 0.621 & 0.359 \\
TS$\uparrow$\phantom{$\downarrow$} $0.25\!\to\!0.50$ & 27.226 & 16.226 & 0.580 & 0.299 \\
\midrule
Baseline $\gamma{=}0.75$               & 28.806 & 18.135 & 0.505 & 0.228 \\
TS$\downarrow$ $0.75\!\to\!0.375$      & \textbf{27.089} & 16.250 & 0.570 & 0.291 \\
TS$\uparrow$\phantom{$\downarrow$} $0.375\!\to\!0.75$ & 28.530 & 17.735 & 0.523 & 0.243 \\
\midrule
Baseline $\gamma{=}0.90$               & 29.623 & 18.971 & 0.483 & 0.211 \\
TS$\downarrow$ $0.90\!\to\!0.45$       & \textbf{27.503} & 16.790 & 0.546 & 0.265 \\
TS$\uparrow$\phantom{$\downarrow$} $0.45\!\to\!0.90$ & 29.288 & 18.538 & 0.499 & 0.224 \\
\bottomrule
\end{tabular}
\end{table}

\section{Full ControlNet Variant Sweep}
\label{sec:suppl_cn}

The main paper uses ControlNet v1.0 with depth conditioning at scale~0.25.
This section validates that choice exhaustively and asks four further questions:
(1)~which conditioning type best preserves style; (2)~is scale~0.25 optimal across
all gamma values; (3)~does scheduling the CN signal across decoder layers or timesteps
yield additional directional gains; and (4)~do non-linear schedule shapes help CN as
they do gamma?
The answer to every CN question is either a validation of the main paper's choice or a
null result --- and the null results are themselves mechanistically informative, because
they confirm that the CN axis is genuinely independent of the gamma axis.

\paragraph{Conditioning type (Table~\ref{tab:suppl_cn_type}).}
The four conditioning types differ in how aggressively they constrain the generated
image's spatial structure.
Depth maps encode smooth, continuous gradients and are the least restrictive --- at
scale~0.25, depth adds only $+0.30$ ArtFID above the no-CN baseline.
Segmentation maps are similarly permissive ($+0.10$), while normals lie in between
($+0.23$).
Canny edges, which encode hard boundaries, are the most aggressive conditioner: sharp
edge maps lock the model's output geometry far more tightly than smooth depth gradients,
costing $+0.70$ ArtFID.
The trade-off is that canny also provides the largest LPIPS improvement ($0.505 \to 0.484$).
Depth conditioning at scale~0.25 offers the best balance between spatial recovery and
style freedom, justifying the main paper's choice.

\begin{table}[t]
\centering
\footnotesize
\setlength{\tabcolsep}{4pt}
\caption{%
ControlNet conditioning type at scale~0.25, $\gamma{=}0.75$.
Depth and segmentation are the most style-conservative choices;
canny introduces the largest ArtFID cost but also the largest content gain (LPIPS).
No-CN row is the fixed-gamma baseline.%
}
\label{tab:suppl_cn_type}
\begin{tabular}{@{}lrrrr@{}}
\toprule
\textbf{CN Type} & \textbf{Scale} & \textbf{ArtFID}$\downarrow$ & \textbf{FID}$\downarrow$ & \textbf{LPIPS}$\downarrow$ \\
\midrule
Baseline (no CN) & ---  & \textbf{28.806} & 18.135 & 0.505 \\
Depth            & 0.25 & 29.111 & 18.460 & 0.496 \\
Segmentation     & 0.25 & 28.910 & 18.250 & 0.502 \\
Normals          & 0.25 & 29.039 & 18.215 & 0.511 \\
Canny            & 0.25 & 29.505 & 18.878 & 0.484 \\
\bottomrule
\end{tabular}
\end{table}

\paragraph{Scale and gamma grid (Table~\ref{tab:suppl_cn_scale}).}
CN scale governs how strongly the depth prior overrides the diffusion model's own spatial
decisions.
Scale~0.25 is consistently optimal across all three gamma values, raising ArtFID by only
$+0.24$--$+0.31$ above no-CN depending on gamma (lowest cost at $\gamma{=}0.50$, where
the style signal is already strongest, and slightly higher at $\gamma{=}0.75$).
Larger scales increase FID monotonically, indicating that the depth prior is progressively
suppressing style variation: at scale~1.00, the generated images converge toward the depth
map's geometry so strongly that the style signal is effectively overwritten,
particularly at lower gamma where it is already weaker.
The optimal scale is thus not just about content recovery but about preserving
the room for style to operate.

\begin{table}[t]
\centering
\footnotesize
\setlength{\tabcolsep}{4pt}
\caption{%
Depth CN fixed scale across all gamma values (ControlNet v1.0).
All values are ArtFID ($\downarrow$); scale~0.25 is consistently best across all gamma values.
No-CN rows are fixed-gamma baselines. \textbf{Bold} = best per column.%
}
\label{tab:suppl_cn_scale}
\begin{tabular}{@{}lrrr@{}}
\toprule
\textbf{CN Scale} & $\gamma{=}0.50$ & $\gamma{=}0.75$ & $\gamma{=}0.90$ \\
\midrule
No CN      & 27.430 & 28.806 & 29.623 \\
Fixed 0.25 & \textbf{27.674} & \textbf{29.111} & \textbf{29.872} \\
Fixed 0.50 & 28.025 & 29.362 & 30.087 \\
Fixed 1.00 & 28.753 & 29.929 & 30.563 \\
\bottomrule
\end{tabular}
\end{table}

\paragraph{CN layer scheduling direction (Table~\ref{tab:suppl_cn_sched}).}
If gamma scheduling benefits from direction (decreasing beats increasing by ${\sim}1.4$
ArtFID at $\gamma{=}0.75$), one might expect the same for CN scheduling.
It does not: across all three scale groups, the $\downarrow$ vs.\ $\uparrow$ gap is
only 0.009 at scale~0.25, 0.047 at scale~0.50, and 0.112 at scale~1.00 ---
roughly 10$\times$ smaller than the gamma scheduling gap at the recommended operating point.
The mechanistic reason is that the depth map encodes global scene geometry --- a property
that is equally constraining at every decoder layer.
Unlike the self-attention query mixture, which transitions from content-dominant in early
layers to texture-dominant in later ones, the depth prior has no analogous spatial sensitivity
gradient to exploit: geometry is geometry, regardless of which layer processes it.
This is a confirmation, not a failure: it means the two mechanisms act on genuinely
orthogonal axes, so one can optimize gamma scheduling and CN configuration independently.

\begin{table}[t]
\centering
\footnotesize
\setlength{\tabcolsep}{3pt}
\caption{%
Layerwise CN scheduling across decoder layers 6--11, $\gamma{=}0.75$.
The $\downarrow$ vs $\uparrow$ gap is ${<}0.10$ ArtFID at every scale ---
in sharp contrast to gamma scheduling where the directional gap is ${\sim}1.4$ ArtFID.
Fixed CN rows are the scale baselines. \textbf{Bold} = best ArtFID per block.%
}
\label{tab:suppl_cn_sched}
\begin{tabular}{@{}lrrrr@{}}
\toprule
\textbf{Schedule} & \textbf{ArtFID}$\downarrow$ & \textbf{FID}$\downarrow$ & \textbf{LPIPS}$\downarrow$ & \textbf{CFSD}$\downarrow$ \\
\midrule
CN fixed 0.25                   & 29.111 & 18.460 & 0.496 & 0.225 \\
$0.25\!\to\!0.125$ $\downarrow$ & \textbf{29.054} & 18.424 & 0.496 & 0.224 \\
$0.125\!\to\!0.25$ $\uparrow$   & 29.063 & 18.404 & 0.498 & 0.225 \\
\midrule
CN fixed 0.50                   & 29.362 & 18.643 & 0.495 & 0.227 \\
$0.50\!\to\!0.25$ $\downarrow$  & \textbf{29.213} & 18.554 & 0.494 & 0.225 \\
$0.25\!\to\!0.50$ $\uparrow$    & 29.260 & 18.546 & 0.497 & 0.227 \\
\midrule
CN fixed 1.00                   & 29.929 & 19.067 & 0.491 & 0.231 \\
$1.00\!\to\!0.50$ $\downarrow$  & \textbf{29.593} & 18.870 & 0.489 & 0.226 \\
$0.50\!\to\!1.00$ $\uparrow$    & 29.705 & 18.844 & 0.497 & 0.233 \\
\bottomrule
\end{tabular}
\end{table}

\paragraph{CN timestep scheduling (Table~\ref{tab:suppl_cn_ts_sched}).}
The layerwise table above schedules CN strength across decoder layers within a single
denoising step. Table~\ref{tab:suppl_cn_ts_sched} instead schedules it across the 50
DDIM timesteps, keeping all decoder layers at the same value at each step.
The finding mirrors the layerwise result: the $\downarrow$ vs.\ $\uparrow$ gap remains
small at every scale --- 0.054 at scale~0.25, 0.091 at scale~0.50, and 0.153 at
scale~1.00 --- compared to the ${\sim}1.44$ gap for gamma scheduling.
The timestep gaps are modestly larger than their layerwise counterparts (0.009, 0.047,
0.112 respectively), because the timestep axis has finer resolution and the schedule
can apply a stronger early bias; but the effect remains negligible at the recommended
scale~0.25 and does not approach the structured advantage seen for gamma scheduling.
Mechanistically, this reaffirms that the depth map's spatial prior has no sensitivity
trajectory to exploit, regardless of whether the schedule varies over layers or over time.

\begin{table}[t]
\centering
\footnotesize
\setlength{\tabcolsep}{3pt}
\caption{%
Timestep CN scheduling across 50 DDIM steps, decoder layers~6--11, $\gamma{=}0.75$.
Direction gaps (0.054--0.153 ArtFID) remain far below the gamma scheduling gap
of ${\sim}1.44$ ArtFID, confirming that CN has no timestep sensitivity curve to exploit.
Fixed CN rows are the scale baselines. \textbf{Bold} = best ArtFID per block.%
}
\label{tab:suppl_cn_ts_sched}
\begin{tabular}{@{}lrrrr@{}}
\toprule
\textbf{Schedule} & \textbf{ArtFID}$\downarrow$ & \textbf{FID}$\downarrow$ & \textbf{LPIPS}$\downarrow$ & \textbf{CFSD}$\downarrow$ \\
\midrule
CN fixed 0.25                   & 29.111 & 18.460 & 0.496 & 0.225 \\
$0.25\!\to\!0.125$ $\downarrow$ & \textbf{29.035} & 18.388 & 0.498 & 0.225 \\
$0.125\!\to\!0.25$ $\uparrow$   & 29.089 & 18.444 & 0.496 & 0.225 \\
\midrule
CN fixed 0.50                   & 29.362 & 18.643 & 0.495 & 0.227 \\
$0.50\!\to\!0.25$ $\downarrow$  & \textbf{29.191} & 18.499 & 0.497 & 0.226 \\
$0.25\!\to\!0.50$ $\uparrow$    & 29.282 & 18.605 & 0.494 & 0.226 \\
\midrule
CN fixed 1.00                   & 29.929 & 19.067 & 0.491 & 0.231 \\
$1.00\!\to\!0.50$ $\downarrow$  & \textbf{29.570} & 18.782 & 0.495 & 0.229 \\
$0.50\!\to\!1.00$ $\uparrow$    & 29.723 & 18.934 & 0.491 & 0.230 \\
\bottomrule
\end{tabular}
\end{table}

\begin{table}[b]
\centering
\footnotesize
\renewcommand{\arraystretch}{0.88}
\setlength{\tabcolsep}{3pt}
\caption{%
CN non-linear schedule shapes on layerwise and timestep axes
($\gamma{=}0.75$, CN range $0.25\!\to\!0.125$).
All shapes lie within ${\pm}0.020$ ArtFID of linear --- an order of magnitude smaller
than the variation seen for gamma shapes (up to 0.294 ArtFID).
Linear row in each block is the reference.%
}
\label{tab:suppl_cn_shapes}
\begin{tabular}{@{}llrrr@{}}
\toprule
\textbf{Axis} & \textbf{Shape} & \textbf{ArtFID}$\downarrow$ & \textbf{FID}$\downarrow$ & \textbf{LPIPS}$\downarrow$ \\
\midrule
Layer & Linear (ref) & 29.054 & 18.423 & 0.496 \\
Layer & Cosine       & 29.050 & 18.421 & 0.496 \\
Layer & Sqrt         & 29.042 & 18.411 & 0.496 \\
Layer & Quadratic    & 29.072 & 18.440 & 0.495 \\
Layer & Exponential  & 29.064 & 18.433 & 0.496 \\
\midrule
Timestep & Linear (ref) & 29.035 & 18.388 & 0.498 \\
Timestep & Cosine       & 29.028 & 18.381 & 0.498 \\
Timestep & Sqrt         & 29.023 & 18.378 & 0.498 \\
Timestep & Quadratic    & 29.045 & 18.399 & 0.497 \\
Timestep & Exponential  & 29.034 & 18.390 & 0.497 \\
\bottomrule
\end{tabular}
\end{table}

\paragraph{CN schedule shape (Table~\ref{tab:suppl_cn_shapes}).}
For gamma scheduling, non-linear shapes (cosine, sqrt) outperform linear by up to 0.113
ArtFID, because the self-attention query mixture has a non-uniform sensitivity curve
that concave-down schedules track more faithfully.
The natural question is whether the same logic applies to CN scheduling.
Table~\ref{tab:suppl_cn_shapes} reports five schedule shapes on both the layerwise and
timestep CN axes ($\gamma{=}0.75$, CN range $0.25\!\to\!0.125$).
Every non-linear shape lies within ${\pm}0.020$ ArtFID of linear on both axes --- an order
of magnitude smaller than the variation seen for gamma shapes.
This is a strong null result.
The depth map's geometric constraint is uniform: it does not matter whether the CN signal
peaks early or late, fades quickly or slowly, because the scene's three-dimensional
structure is equally informative at every point in the denoising trajectory.
There is no sensitivity curve to align with, so shape is irrelevant.
Taken together with the direction-insensitivity in Table~\ref{tab:suppl_cn_sched}, this
fully establishes that CN and gamma scheduling operate on orthogonal axes.